\documentclass[runningheads,a4paper]{llncs}

\usepackage{amssymb}
\usepackage{graphicx}

\newcommand{\keywords}[1]{\par\addvspace\baselineskip
\noindent\keywordname\enspace\ignorespaces#1}

\begin{document}

\mainmatter  

\title{Uncertainty in Ontology Matching: a Decision Rule-based Approach}

\titlerunning{Uncertainty in Ontology Matching: a Decision Rule-based Approach}

%
%
\author{Amira Essaid \inst{1,2}%
\and Arnaud Martin \inst{2}\and Gr\'egory Smits \inst{2} \and Boutheina Ben Yaghlane\inst{3}\\
}
\authorrunning{Amira Essaid et al.}

\institute{${}^1$ LARODEC, University of Tunis, ISG Tunis, Tunisia\\
${}^2$IRISA, University of Rennes1, Lannion, France\\
${}^3$ LARODEC, University of Carthage, IHEC Carthage, Tunisia} 

%
%

\maketitle

\begin{abstract}
Considering the high heterogeneity of the ontologies published on the web, ontology matching is a crucial issue whose aim is to establish links between an entity of a source ontology and one or several entities from a target ontology. Perfectible similarity measures, considered as sources of information, are combined to establish these links. The theory of belief functions is a powerful mathematical tool for combining such uncertain information. In this paper, we introduce a decision process based on a distance measure to identify the best possible matching entities for a given source entity.

\end{abstract}

\section{Introduction}

This paper proposes a decision rule based on a distance measure. This rule calculates the distance between a combined mass function and a categorical mass function and keep the hypotheses with the lowest distance. We propose this rule for its ability to give decision on composite hypotheses as well as its convenience to our domain of application, namely the semantic web and particularly the ontology matching where decision making is an important step.

Ontology matching is the process of finding for each entity of a source ontology $O_1$ its corresponding entity in a target ontology $O_2$. This process can focus on finding simple mappings (1:1) or complex mappings (1:n or n:1). The first consists in matching only one entity of $O_1$ with only one entity of $O_2$ whereas the second consists in finding either for one entity of $O_1$ its multiple correspondences of entities in $O_2$ or matching multiple entities of $O_1$ with only one entity of $O_2$. We are interested in this paper in finding simple mappings as well as the complex one of the form (1:n).

The matching process is performed through the application of matching techniques which are mainly based on the use of similarity measures. Since no similarity measure applied individually is able to give a perfect alignment, the exploitation of the complementarity of different similarity measures can yield to a better alignment. Combining these similarity measures may raise conflicts between the different results which should be modeled and resolved. 

We suggest to use the theory of belief functions \cite{dempster,shafer} as a tool for modeling the ontology matching and especially for combining the results of the different similarity measures. Due to the fact that we are working on an uncertain aspect and we are interested in finding complex matching which can be viewed as finding composite hypotheses formed from entities of two ontologies, we suggest to apply our proposed decision rule on the combined information and to choose for each entity of the source ontology, the entities of the target ontology with the lowest distance.

The remainder of this paper is organized as follows: we are interested in section 2 in defining the ontology matching process. In section 3, we recall the basic concepts underlying the theory of belief functions. In section 4, we present our decision rule based on a distance measure. Section 5 is devoted to the description of the credibilistic decision process for matching ontologies as well as the application of our proposed decision rule. Section 6 discusses an overview of some ontology matching approaches dealing with uncertainty. Finally, we conclude in section 7 and present future work.

\section {Ontology Matching}

The open nature of the semantic web \cite{berners} tends to encourage the development, for a domain of interest, of heterogeneous ontologies which differ from each other at the terminological level and/or the representational one. In order to mitigate the effect of semantic heterogeneity and to assure interoperability between applications that make use of these ontologies, a key challenge is to define an efficient and reliable \textit{matching between ontologies} \cite{euzenat}.

Formally, ontology matching is defined as a function A = f($O_1$, $O_2$, A', p, r).
In fact, from a pair of ontologies to match \textit{$O_1$} and \textit{$O_2$}, an input alignment \textit{A'}, a set of parameters \textit{p}, a set of oracles and resources \textit{r}, the function \textit{f} returns an alignment \textit{A} between these ontologies.
We note that parameters and resources refer to thresholds and external resources respectively.


With the new vision of the web that tends to make applications understandable by machines, an automatic and semi automatic discovery of correspondences between ontologies is required. The reader may refer to \cite{euzenat} for an exhaustive state of the art of ontology matching techniques. 

%

\section{The Theory of Belief Functions}

\subsection{Definitions}

The frame of discernment \textit{$\Theta$}  = $\left\{ \theta_1, \theta_2, \ldots, \theta_n \right\}$ is a finite non empty set of \textit{n} elementary and mutually exclusive and exhaustive hypotheses related to a given problem. The power set of \textit{$\Theta$}, denoted by \textit{$2^{\Theta}$} is defined as the set of singleton
hypotheses of $\Theta$, all possible disjunctions of these hypotheses as well as the empty set.

The basic belief assignment (\textit{bba}) is the mapping from elements of the power set \textit{$2^{\Theta}$} onto $\left[0, 1\right]$ that satisfies:
\begin{equation}
m(\emptyset) = 0  ,   \sum_{A \subseteq \Theta} m(A)=1.
\end{equation}

The value \textit{m(A)} quantifies the part of belief exactly committed to the subset A of \textit{$\Theta$}.

A focal element A is an element of \textit{$2^{\Theta}$} such that $m(A)\neq 0$.

From a given bba , the corresponding credibility and plausibility functions are respectively defined as:
\begin{equation}
\begin{tabular}{l}
$bel(A) = \displaystyle{\sum_{B\subseteq A, B\neq\emptyset}}m(B)$.\\
\end{tabular}
\end{equation}

and 
\begin{equation}
\begin{tabular}{l}
$pl(A) = \displaystyle{\sum_{A\cap B\neq\emptyset}}m(B)$.\\
\end{tabular}
\end{equation}

The value \textit{bel(A)} expresses the total belief that one allocates to A whereas the \textit{pl(A)} quantifies the maximum amount of belief that might support a subset A of $\Theta$.

Some special \textit{bbas} are defined in the theory of belief functions. Among them, the categorical bba which is a bba with a unique focal element different from the frame of discernment $\Theta$ and the empty set $\emptyset$, and which is defined as $m_X(X) =1$. 
	
	\subsection{Combination of Belief Functions}
Let $S_1$ and $S_2$ be two distinct and independent sources providing two different bbas $m_1$ and $m_2$ defined on the same frame of discernment $\Theta$. These two bbas are combined by either the conjunctive rule of combination or the disjunctive rule.
\begin{itemize}
	\item The conjunctive rule  of combination is used when the two sources are fully reliable. This rule is defined in \cite{smets} as :
	\begin{equation}
		\label{SmetsRule}
			m_{1\textcircled{\scriptsize{$\cap$}}2}(A)=\sum _{B\cap C=A} m_1(B)\times m_2(C).
	\end{equation}
	The conjunctive rule can be seen as an unnormalized Dempster's rule of combination \cite{dempster} which is defined by:
	\begin{equation}
m_{1\oplus 2}(A)=
\left\{
\begin{tabular}{ll}
$\frac{\displaystyle{\sum _{B\cap C=A}} m_1(B)\times m_2(C)}{1-\displaystyle{\sum _{B\cap C=\emptyset}} m_1(B)\times 
m _2(C)}$ & $\forall A\subseteq\Theta,\hspace{0.1cm}A\neq \emptyset$\\
0 & $ if \hspace{0.1cm}A=\emptyset$\\
\end{tabular}
\right.
\end{equation}

The Dempster's rule of combination is normalized through $1-\displaystyle{\sum _{B\cap C=\emptyset}} m_1(B)\times m _2(C)$ and it works under the closed world assumption where all the possible hypotheses of the studied problem are supposed to be enumerated on $\Theta$.
	
	\item The disjunctive rule is used when at least one of the sources is reliable without knowing which one of them. It is defined in \cite{smets} by:
	\begin{equation}
m_{1\textcircled{\scriptsize{$\cup$}}2}(A)=\sum _{B\cup C=A} m_1(B)\times m_2(C).
\end{equation}
\end{itemize}

\subsection{Decision Making}
Combining information provided by the different sources leads to a global one that has to be analyzed in order to choose the most likely hypothesis. Decision making can be held in two different ways. 
\begin{itemize}
	\item Decision on singletons:  It means that the most likely solution to a given problem is one of the hypothesis of $\Theta$. To determine this most likely solution, one may:
					\begin{itemize}
						\item Maximize the credibility: It consists on retaining the most credible hypothesis by giving the minimum of chances to each of the disjunctions.
						\item Maximize the plausibility: It consists on retaining the most plausible hypothesis by giving the maximum of chances to each of the singletons.
						\item Maximize the pignistic probability: It was introduced in \cite{smets05} and it is the common used decision function because it represents a compromise between the maximum of credibility and the maximum of plausibility. The pignistic probability consists on choosing the most probable singleton hypothesis by dividing the mass attributed to each hypothesis, different from the singleton hypothesis, by the hypotheses composing it. It is given for each $A \in 2^{\Theta}$, $A \neq \emptyset$ by:
							\begin{equation}
								\begin{tabular}{ll}

								$\displaystyle{betP(X) = \sum_{A \in 2^\Theta, X \in A}\frac{m(A)}{|A|(1-m(\emptyset))}}$.
								\end{tabular}
								\end{equation}
	where $|A|$ represents the cardinality of A.
					\end{itemize}
	\item Decision on unions of singletons: Few works were interested in making decision on composite hypotheses (\cite{denoeux}, \cite{appriou}, \cite{martin}). The  approach proposed in \cite{appriou} helps to choose a solution of a given problem by considering all the elements contained in $2^\Theta$. This approach weights the decision functions listed previously by an utility function depending on the cardinality of the elements. For each A $\in 2^\Theta$ we have: 
		\begin{equation}
			\displaystyle{A= \mathop{argmax} _{X\in 2^\Theta} (m_d(X)pl(X))}
		\end{equation}
		where $m_d$ is a mass defined by: 
		\begin{equation}
			m_d(X) = K_d\lambda_{X}(\frac{1}{|X|^r})
		\end{equation}

\end{itemize}
\textit{r} is a parameter in $\left[0, 1\right]$ for choosing a decision. When \textit{r }is equal to 0 it reflects a total indecision and when it is equal to 1 it means that we decide on a singleton. The value $\lambda_{X}$ is used to integrate the lack of knowledge about one of the elements X of \textit{$2^\Theta$}. $K_d$ is a normalization factor. 

\section {Decision Rule Based on a Distance Measure}

We aim in this paper to propose a decision rule helping us to choose the most likely hypothesis for a given problem after combining the information provided by different sources of information, i.e. bbas. This rule, based on a distance measure, is inspired from \cite{smarandache} and is defined as:

\begin{equation}
	A = argmin(d(m, m_X))
\end{equation}
The proposed rule aims at calculating the distance between \textit{m} which is a combined bba (obtained after applying a combination rule) and $m_X$ is the categorical bba of X such that X $\in$ $2^\Theta$.
The most likely hypothesis to choose is the hypothesis whose categorical bba is the nearest to the combined bba.\\

In order to make a decision:
\begin{itemize}
	\item First, we have to identify the elements for which we have to construct the categorical bba. In fact, we choose to work on elements of $2^\Theta$ such that the cardinality of the element is less or equal to 2. This filtering is due to the fact that we want to limit the number of elements to be considered especially with a power set $2^\Theta$ of large cardinality.
	\item Then, we construct the categorical bba for each of the selected element.
	\item Finally, we calculate the distance between the combined bba and each of the categorical bbas. The minimum distance is kept and our decision corresponds to the categorical bba's element having the lowest distance with the combined bba.
\end{itemize}

For the calculation of the distance between the bbas, we use the Jousselme distance \cite{jousselme} which is specific to the theory of belief functions because of the matrix D defined on $2^\Theta$. This distance has the advantage to take into account the cardinality of the focal elements. This distance is defined for two bbas $m_1$ and $m_2$ as follows:
\begin{equation}
d(m_1,m_2)=\sqrt{\frac{1}{2}(m_1-m_2)^t\underline{\underline{D}}(m_1-m_2)}
\end{equation}
where \underline{\underline{D}} is a matrix based on Jaccard distance as a similarity measure between focal elements. This matrix is defined as:
\begin{equation}
D(A,B)=\left\{
\begin{tabular}{ll}
1&if A=B=$\emptyset$\\
$\frac{\mid A\cap B\mid}{\mid A\cup B\mid}$&$\forall A,B \in 2^\Theta$\\
\end{tabular}
\right.
\end{equation}

To illustrate the proposed decision rule, we take the following example. Let's consider the frame of discernment $\Theta = \left\{\theta_1, \theta_2, \theta_3 \right\}$. The list of elements for which we have to construct their categorical bba are $\left\{\theta_1, \theta_2, \theta_3,\theta_1\cup\theta_2, \theta_1\cup\theta_3, \theta_2\cup\theta_3 \right\}$.
Suppose that we have two sources $S_1$ and $S_2$ providing two different bbas $m_1$ and $m_2$ defined on the frame of discernment $\Theta$. The table 1 illustrates these two bbas as well as their combined bba obtained after applying the Dempster's rule of combination.
\begin{table}
\caption{bba1 and bba2 and their combined bba}
\begin{center}
\begin{small}
\begin{tabular}{|c|c|c|} \hline
        bba1                                   &      bba2           & combined bba \\ \hline
        $m_1(\theta_1)$ = 0.4                &   $m_2(\theta_2)$ = 0.2   &$m_{comb}(\theta_1)$ = 0.3478\\ 
        $m_1(\theta_2 \cup \theta_3)$ = 0.2  &   $m_2(\Theta)$ = 0.8     &$m_{comb}(\theta_2)$ = 0.1304 \\
        $m_1(\Theta)$ = 0.4                  &                         &$m_{comb}(\Theta)$ = 0.3478\\
									                         &                     &$m_{comb}(\theta_2 \cup \theta_3)$ = 0.1739 \\\hline
\end{tabular}
\end{small}
\end{center}
\label{bbas}
\end{table}

The application of our proposed decision rule gives the results illustrated in table 2 where it shows for every element the distance obtained between the categorical bba of this element and the combined bba.
\begin{table}
\caption{Results of the proposed decision rule}
\begin{center}
\begin{small}
\begin{tabular}{|c|c|c|} \hline
        Element                 & Distance \\ \hline
        $\theta_1$              &  0.537\\ 
        $\theta_2$              &  0.647\\
        $\theta_3$              &  0.741\\
				$\theta_1\cup\theta_2$  &  0.472\\
				$\theta_1\cup\theta_3$  &	 0.536\\
				$\theta_2\cup\theta_3$  &  0.529\\
				\hline
\end{tabular}
\end{small}
\end{center}
\label{bbas}
\end{table}

Based on the results obtained in table 2, the most likely hypothesis to choose is the element $\theta_1\cup\theta_2$.

\section {Credibilistic Decision Process for Ontology Matching}

In \cite{essaid}, we proposed a credibilistic decision process for ontology matching. In the following, we describe this process occurring mainly in three steps, then we will apply the proposed decision rule in order to find a correspondence for a given entity of the source ontology.

\subsubsection{1- Matching ontologies:}

We apply three name-based techniques (Levenshtein distance, Jaro distance and Hamming distance) for matching two ontologies $O_1$ and $O_2$ related to conference organization\footnote {http ://oaei.ontologymatching.org/2013/conference/index.html}. We have the following results:
\begin{table}[htbp]
\caption{Results of matching the entity ConferenceMember of $O_1$ with entities of $O_2$}
\begin{center}
\begin{small}
\begin{tabular}{|c|c|c|} \hline
        method                 &       $e_2 \in O_2$&$n$              \\ \hline
        Levenshtein & $Conference\_fees$&   0.687            \\
        Jaro&$Conference$&0.516 \\ 
        Hamming& $Conference$&0.625\\\hline
\end{tabular}
\end{small}
\end{center}
\label{resultatMapping}
\end{table}

This table shows that using the levenshtein distance, the entity \textit{ConferenceMember} matches to \textit{Conference\_fees} with a confidence value of 0.687.

\subsubsection{2- Modeling the matching under the theory of belief functions:}
We are interested here in modeling the matching results obtained in the previous step under the theory of belief functions.
\begin{itemize}
	\item \textit{Frame of discernment}: It contains all the entities of the target ontology $O_2$ for which a corresponding entity in the source ontology $O_1$ exists.
	\item \textit{Source of information}: Every correspondence established by one of the matching techniques is considered as an information given by a source.
	\item \textit{Basic Belief Assignments (bba)}: Once we get all the correspondences, we keep only those where an entity source $e_1 \in O_1$ has a correspondence when applying the three techniques. Then, we construct for each of the selected correspondence its mass function. The similarity measure obtained after applying a matching technique is interpreted as a mass. Due to the fact that for a source of information, the sum of mass functions has to be equal to 1, a mass will be affected to the total ignorance.
	Let's take the results illustrated in Table 3. In this table, we have information provided by three different sources respectively denoted by $S_{lev}^{e_1}$, $S_{jars}^{e_1}$ and $S_{hamming}^{e_1}$, where $e_1 = ConferenceMember$. The bba related to the source $S_{lev}^{e_1}$ is: $m_{S_{lev}^{e_1}}(Conference\_fees) = 0.687$ and $m_{S_{lev}^{e_1}}(\Theta) = 1-0.687 = 0.313$. The bbas for the other sources are constructed in the same manner.
	\item \textit{Combination}: Let's resume the obtained bbas of the three sources. We have:
	\begin{itemize}
	\item $m_{S_{lev}^{e_1}}(Conference\_fees) = 0.687$ and $m_{S_{lev}^{e_1}}(\Theta) = 0.313$	
		\item $m_{S_{jaro}^{e_1}}(Conference) = 0.516$ and $m_{S_{jaro}^{e_1}}(\Theta) = 0.484$
		\item $m_{S_{hamming}^{e_1}}(Conference) = 0.625$ and $m_{S_{hamming}^{e_1}}(\Theta) = 0.375$
	\end{itemize}
	Once we apply the Dempster's rule of combination, we obtain the following results:
	\begin{itemize}
	\item	$m_{comb}^{e_1}(Conference\_fees) = 0.2849$
	\item $m_{comb}^{e_1}(Conference) = 0.5853$
	\item $m_{comb}^{e_1}(\Theta) = 0.1298$
	\end{itemize}
	
\end{itemize}
\subsubsection{3- Decision Making:} Based on the combined bba which takes into account all the information provided by the different sources, we will be able in this step to choose for each entity of the source ontology its corresponding in the target ontology. For example, for the entity \textit{ConferenceMember}, we will be able to decide if we have to match it with \textit{Conference\_fees }or \textit{Conference} or simply we will not have a precise decision but rather an uncertain one where we can match \textit{ConferenceMember} to $Conference\_fees\cup Conference$. We are interested in our credibilistic process to get an uncertain decision. For this purpose, we apply our proposed decision rule. First, we construct the categorical bba of elements having a cardinality equal to 2. For the example illustrated in figure 1 we have:
\begin{itemize}
	\item m($conference\_document \cup conference$) = 1
	\item m($conference \cup conference\_fees$) = 1
	\item m($conference\_volume \cup committee$) = 1
	\item  \ldots
\end{itemize}

Then we calculate the distance between the combined bba obtained previously and each of the categorical bba. Our best alignment corresponds to the nearest element to the combined bba in other words the element whose categorical bba has the minimum distance with the combined bba. For the entity \textit{ConferenceMember} of the ontology $O_1$ we find $conference\_fees \cup conference$ with a distance equal to 0.52. This process is repeated for each entity of the source ontology in order to identify the most significant correspondences in the target ontology.

\section{Related Works}

Only few ontology matching methods have considered that dealing with uncertainty in a matching process is a crucial issue. We are interested in this section to present some of them where the probability theory \cite{pan} and the Dempster-Shafer theory (\cite{besana},\cite{nagy},\cite{wang}) are the main mathematical models used.
In \cite{pan}, the authors proposed an approach for matching ontologies based on bayesian networks which is an extension of the BayesOWL. The BayesOWL consists in translating an OWL ontology into a bayesian network (BN) through the application of a set of rules and procedures. In order to match two ontologies, first the source and target ontologies are translated into $BN_1$ and $BN_2$ respectively. The mapping is processed between the two ontologies as an evidential reasoning between $BN_1$ and $BN_2$. The authors assume that the similarity information between a concept $C_1$ from a source ontology and a concept $C_2$ from a target ontology is measured by the joint probability distribution P($C_1$, $C_2$).

In \cite{besana}, the author viewed ontology matching as a decision making process that must be handled under uncertainty. He presented a generic framework that uses Dempster-Shafer theory as a mathematical model for representing uncertain mappings as well as combining the results of the different matchers. Given two ontologies $O_1$ and $O_2$, the frame of discernment represents the Cartesian product e x $O_2$ where each hypothesis is the couple $<e, e_i>$ such as e $\in O_1$ and $e_i \in O_2$. Each matcher is considered as an expert that returns a similarity measure converted into a basic belief mass. The Dempster rule of combination is used to combine the results provided by a matcher. The pairs with plausibility and belief below a given threshold are discarded. The remaining pairs represent the best mapping for a given entity.

Although, the authors in \cite{nagy} handle uncertainty in the matching process, their proposal differ from that proposed in \cite{besana}. In fact, they use the Dempster-Shafer theory in a specific context of question answering where including uncertainty may yield to better results. Not like in \cite{besana}, they did not give in depth how the frame of discernment is constructed. In addition to that, uncertainty is handled only once the matching is processed. In fact, the similarity matrix is constructed for each matcher. Based on this matrix, the results are modeled using the theory of belief functions and then they are combined.

In \cite{wang}, the authors focused on integrating uncertainty when matching ontologies. The proposed method modeled and combined the outputs of three ontology matchers. For an entity e $\in O_1$, the frame of discernment $\Theta$ is composed of mappings between e and all the concepts in an ontology $O_2$. The different similarity values obtained through the application of the three matchers are interpreted as mass values. Then, a combination of the results of the three matchers is performed.

\section{Conclusion and Perspectives}
In this paper, we proposed a decision rule based on a distance measure. This decision rule helps to choose the most likely hypothesis for a given problem. It is based on the calculation of the distance between a combined bba and a categorical bba. We apply this rule in our proposed credibilistic decision process for the ontology matching. First, we match two ontologies. Then, the obtained correspondences are modeled under the theory of belief functions. Based on the obtained results, a decision making is performed by applying our proposed decision rule.

In the future, we aim at applying other matching techniques. We are interested also in constructing an uncertain ontology based on the obtained results after a decision making and handling experimentations to qualitatively assess the relevance of our approach.

%
%
%
%
%
\end{document}